\documentclass[]{spie}  

\usepackage[table]{xcolor}
\usepackage{amssymb}
\usepackage{pifont}

\usepackage{amsmath,amsfonts,amssymb}
\usepackage{subcaption}
\usepackage{graphicx}
\usepackage{cite} 
\usepackage{times}
\usepackage{epsfig}
\usepackage{amsmath}
\usepackage{nccmath}
\usepackage{amssymb}
\usepackage{mwe}
\usepackage{acro}
\usepackage{amssymb}
\usepackage{xcolor,colortbl}
\usepackage{tabularx}
\usepackage{relsize}
\usepackage{pifont}
\usepackage{booktabs} 
\usepackage{multirow}
\usepackage{multicol}
\usepackage{adjustbox}
\usepackage{float}
\usepackage{graphicx}
\usepackage{makecell}
\usepackage{tabu}
\usepackage[colorlinks=true, allcolors=blue]{hyperref}
\usepackage[capitalize]{cleveref}

\title{CoFi: A Fast Coarse-to-Fine Few-Shot Pipeline for Glomerular Basement Membrane Segmentation}

\author[a]{Hongjin Fang}
\author[b]{Daniel Reisenbüchler}
\author[c]{Kenji Ikemura}
\author[a,c,d]{Mert R. Sabuncu}
\author[c,e]{Yihe Yang}
\author[c]{Ruining Deng}

\affil[a]{Cornell University, Ithaca, NY 14853, USA}
\affil[b]{University of Regensburg, Regensburg, Bavaria 93053, DE}
\affil[c]{Weill Cornell Medicine, New York, NY 10065, USA}
\affil[d]{Cornell Tech, New York, NY 10044, USA}
\affil[e]{Northwell Health, New Hyde Park, NY 11040, USA}

\begin{document} 
\maketitle

\begin{abstract} 
Accurate segmentation of the glomerular basement membrane (GBM) in electron microscopy (EM) images is fundamental for quantifying membrane thickness and supporting the diagnosis of various kidney diseases. While supervised deep learning approaches achieve high segmentation accuracy, their reliance on extensive pixel-level annotation renders them impractical for clinical workflows. Few-shot learning can reduce this annotation burden but often struggles to capture the fine structural details necessary for GBM analysis.
In this study, we introduce CoFi, a fast and efficient coarse-to-fine few-shot segmentation pipeline designed for GBM delineation in EM images. CoFi first trains a lightweight neural network using only three annotated images to produce an initial coarse segmentation mask. This mask is then automatically processed to generate high-quality point prompts with morphology-aware pruning, which are subsequently used to guide SAM in refining the segmentation. The proposed method achieved exceptional GBM segmentation performance, with a Dice coefficient of 74.54\% and an inference speed of 1.9 FPS. We demonstrate that CoFi not only alleviates the annotation and computational burdens associated with conventional methods, but also achieves accurate and reliable segmentation results. The pipeline’s speed and annotation efficiency make it well-suited for research and hold strong potential for clinical applications in renal pathology. The pipeline is publicly available at: https://github.com/ddrrnn123/CoFi.

\end{abstract}

\keywords{Segmentation; Few-Shot Learning; Glomerular Basement Membrane; Segment Anything Model; Prompt engineering}

\section{Introduction}  
\label{sec:intro}

Precise segmentation of the glomerular basement membrane (GBM) in electron microscopy (EM) images is crucial for accurately measuring membrane thickness and diagnosing a wide spectrum of kidney diseases~\cite{Li2020deep,Nyengaard1992,Pichler2004}. Conditions such as Alport syndrome~\cite{Hudson2003} and diabetic nephropathy~\cite{Ota1995} rely on these measurements for reliable clinical assessment. Although supervised learning methods have set a high standard for GBM segmentation~\cite{Chen2021,He2020,Barisoni2020}, they come with the steep cost of requiring large amounts of pixel-level annotated data, which is both time-consuming and impractical in most clinical environments~\cite{Litjens2017}.

\begin{figure}[htbp]
  \centering
  \includegraphics[width=0.8\textwidth]{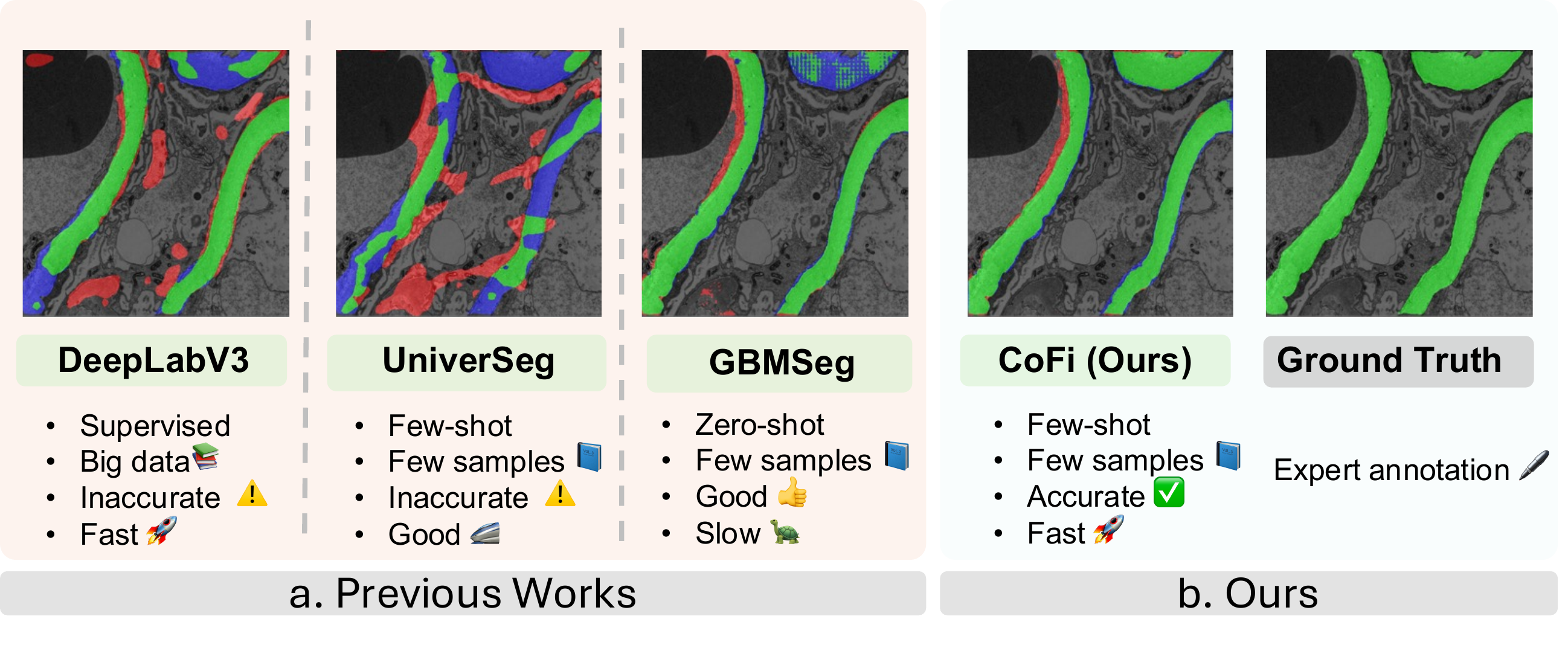}
  \caption{Comparison of GBM segmentation across prior methods and our proposed CoFi pipeline. DeepLabV3 requires large datasets yet remains inaccurate; UniverSeg works with few samples but is inconsistent; and GBMSeg is accurate but slow. CoFi delivers few-shot, accurate, and fast segmentation, closely matching expert annotations.}
  \label{fig:initial}
\end{figure}

Few-shot learning has emerged as a promising alternative, aiming to reduce this annotation burden by training models with just a handful of labeled examples~\cite{Snell2017,Shaban2017,Wang2020}. However, in practice, these approaches often fall short when applied directly to GBM segmentation, as they struggle to capture the subtle and complex ultrastructural details needed by pathologists. Meanwhile, the Segment Anything Model (SAM) has shown remarkable potential for zero-shot segmentation across various domains, provided it receives thoughtfully selected point prompts~\cite{deng2025segment, cui2024all, wang2024sammed,li2024leverage}.However, effectively leveraging SAM to its full potential poses considerable challenges: generating high-quality prompts usually requires loading multiple large pre-trained models and painstakingly adjusting points, a process that introduces significant computational load and delays~\cite{zhou2023can,liu2023prompt,singh2023prompt}.

To address the challenges of significant computational burden and required manual tuning, we present CoFi, a fast and efficient coarse-to-fine few-shot pipeline for GBM segmentation. Our approach begins by training a lightweight neural network on just three annotated images to generate an initial, coarse segmentation mask. This mask is then automatically transformed into high-quality point prompts with morphology-aware pruning, which guide SAM 2~\cite{cheng2023sam2}. SAM 2 is a prompt-learning extension of SAM that enables efficient and adaptive refinement, producing finely detailed and useful segmentations. By combining the efficiency of few-shot learning with automated prompt generation, CoFi delivers accurate results at speeds suitable for research and hold strong potential for integration into real-world clinical workflows. The pipeline is publicly available at: https://github.com/ddrrnn123/CoFi.

\section{Method}

\begin{figure}[htbp]
  \centering
  \includegraphics[width=1.0\textwidth]{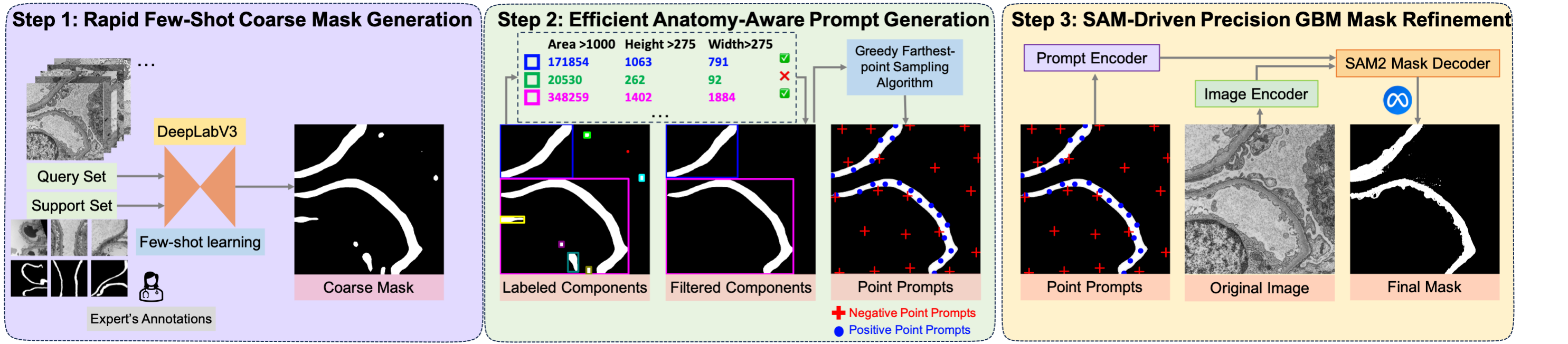}
  \caption{Workflow of CoFi, the fast coarse-to-fine few-shot pipeline for GBM segmentation. Step 1 generates coarse GBM masks using few-shot DeepLabV3. Step 2 extracts anatomy-aware positive and negative point prompts. Step 3 refines the segmentation with SAM 2 to produce the final detailed mask.}
  \label{fig:workflow}
\end{figure}

We introduce a three-step few-shot segmentation pipeline for accurate and efficient delineation of GBM from grayscale TEM images, as illustrated in Figure~\ref{fig:workflow}. The pipeline comprises the following components: (1) Rapid Few-Shot Coarse GBM Mask Generation, (2) Efficient Anatomy-Aware Prompt Generation, and (3) SAM-Driven Precision GBM Mask Refinement.

\subsection{Step 1: Rapid Few-Shot Coarse GBM Mask Generation}

Recent advances in supervised deep learning have produced a range of end-to-end segmentation models, such as U-Net~\cite{ronneberger2015u}, Swin UNETR~\cite{hatamizadeh2022swin}, and DeepLabV3~\cite{chen2017rethinking}, that have demonstrated significant success in biomedical image analysis due to their strong representational capacity and rapid inference speed. Nevertheless, these models are fundamentally reliant on large annotated datasets for optimal performance. In few-shot settings, especially when adapting to new domains with limited supervision, their segmentation quality diminishes substantially, thereby restricting their practicality for clinical GBM analysis. While meta-learned few-shot models like UniverSeg~\cite{valvano2023universeg} show the potential for cross-task generalization, they often fall short in providing the structural precision essential for delineating the highly detailed GBM, where subpixel boundary accuracy is required. These performance gaps are exemplified in the qualitative comparison shown in Figure~\ref{fig:initial}.

To address these challenges, our pipeline first leverages the rapid inference capabilities of end-to-end models to generate an initial coarse segmentation mask, which forms the basis for further refinement. Among the evaluated models, DeepLabV3 was selected for its superior initial localization quality in GBM segmentation as shown in Table~\ref{tab:segmentation_performance}. Unlike other few-shot methods that require extensive computation with large-scale pretrained models, our pipeline utilizes a lightweight, trainable segmentation backbone to efficiently produce robust coarse masks from only a small set of labeled images.

\subsection{Step 2: Efficient Anatomy-Aware Prompt Generation}

While the initial coarse masks offer valuable localization cues, they often lack the precision required for clinical use. As shown in Step 1 of Figure~\ref{fig:workflow}, the coarse mask is inaccurate, underscoring the need for an effective refinement strategy. State-of-the-art models like SAM 2~\cite{cheng2023sam2} offer robust zero-shot segmentation capabilities, contingent on the quality of point prompts provided. However, current methods such as GBMSeg~\cite{yang2024gbmseg} rely on dense patch-wise feature matching with DINOv2~\cite{oquab2023dinov2} to identify positive and negative regions, which is a computationally intensive process that is impractical for real-time or high-throughput clinical settings.

To address the limitations of conventional prompt selection, we introduce a computationally efficient, anatomy-aware algorithm for generating and selecting high-quality point prompts based on GBM morphological characteristics. This process begins with connected component analysis of a coarse GBM mask, followed by Morphology-Aware Pruning, where candidate regions are quantitatively filtered by area, height, and width to retain only those consistent with known GBM anatomical profiles. This targeted filtering suppresses noise and artifacts while enhancing the biological relevance of the selected components. Subsequently, a greedy farthest-point sampling algorithm is used to extract spatially diverse foreground and background points from these refined regions, ensuring that the prompts robustly capture both the structural complexity and semantic context of the GBM. Together, these steps enable efficient and anatomically meaningful prompt generation, directly supporting precise and reliable downstream segmentation refinement.

\subsection{Step 3: SAM-Driven Precision GBM Mask Refinement}

In the final stage, the selected positive and negative point prompts are provided to SAM 2, which generates refined segmentation masks. Its advanced prompt-based zero-shot framework enhances boundary delineation, preserves the integrity of connected components, and minimizes artifacts. This yields viable segmentation outputs with minimal supervision, consistently outperforming the initial coarse masks in both accuracy and reliability.

\section{Data \& Experiments}  
\subsection{Data}  
A total of 196 medical kidney biopsy samples received and reviewed at Northwell Health between January 2023 and July 2023 were retrospectively included in this study. Kidney biopsy tissues were fixed with 2--2.5\% glutaraldehyde in 0.1\,M phosphate buffer (pH 7.0--7.3) and examined using a JEOL JEM-100CX II transmission electron microscope with a tungsten filament at 100\,kV. Digital images were acquired with an Advanced Microscopy Techniques (AMT) camera (model: 1412AM-T1-FW-AM, Woburn, MA). Ultrathin sections were examined using standard techniques. De-identified EM images of glomeruli were analyzed, and glomerular basement membranes were manually annotated in QuPath.

\subsection{Experiment Setting} 
The DeepLabV3 with ResNet-50 backbone was trained for 60 epochs with a batch size of 1, following the settings used in GBMSeg and using the Adam optimizer and a learning rate of $1 \times 10^{-4}$. A composite loss function combining Binary Cross-Entropy with logits (weighted at 0.7) and soft Dice loss (weighted at 0.3) was used to optimize performance. These parameters were selected through empirical tuning to balance convergence speed and segmentation accuracy. The model was initialized with pretrained weights, and Batch Normalization layers were frozen during training to preserve stable feature statistics. Data augmentations included random horizontal flips, affine transformations (shift, scale, rotate), and normalization with a mean and standard deviation of 0.5. Masks were binarized by setting all non-zero pixel values to 1. Five separate training runs with different randomly selected support sets were performed to evaluate model generalization robustness. Results were averaged across these runs.

\subsection{Inference and Post-Processing}
During inference, the trained DeepLabV3 model generates a coarse GBM mask that undergoes connected component analysis and extent-based filtering to isolate reliable regions. Connected component filters components of sizes smaller than 1000 pixels, removing small, noisy predictions, while extent-based filtering discards low-confidence structures smaller than 275 pixels in width or height. Following this refinement, a total of 40 point prompts (20 positive and 20 negative) are sampled per image using a farthest-point sampling strategy across high- and low-confidence regions. These point prompts are then passed into SAM2 for final mask prediction. This postprocessing step enables refined delineation of the GBM boundary with high accuracy and minimal latency.

\subsection{Evaluation Metrics}
Model performance was evaluated using standard segmentation metrics, including the Dice coefficient score (DSC), Intersection over Union (IoU), and pixel-level accuracy (ACC). In addition to these accuracy metrics, we also measured inference speed in frames per second (FPS), defined as the number of full-resolution images processed per second during evaluation. This metric reflects the system’s computational efficiency and its suitability for time-sensitive clinical applications.

\begin{figure}[htbp]
  \centering
  \includegraphics[width=1.033\textwidth]{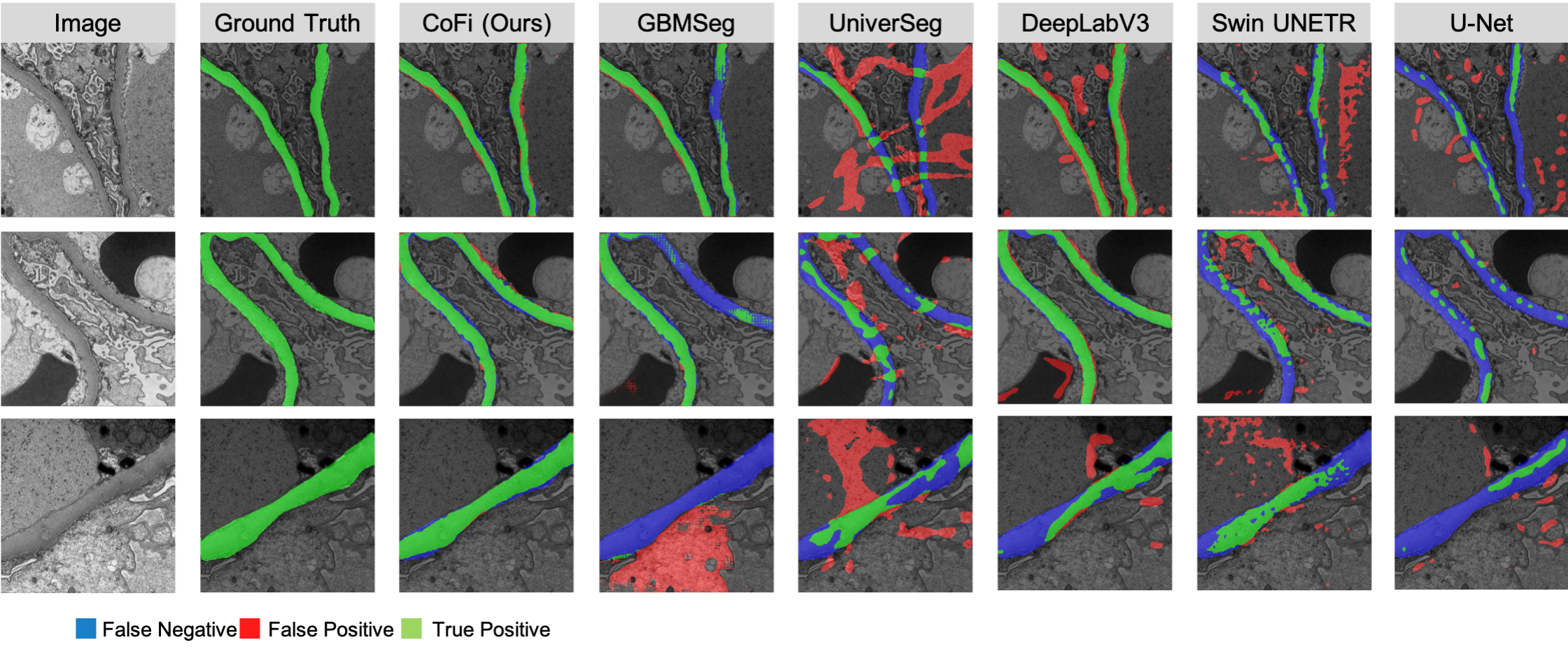}
  \caption{Qualitative comparison of GBM segmentation on a representative test image. Overlaid colored markers denote true positives (green), false positives (red), and false negatives (blue), demonstrating CoFi’s closer alignment with the ground truth and fewer segmentation errors compared to existing methods.}
  \label{fig:qualitative_comp}
\end{figure}

\section{Results} 
We evaluated the proposed CoFi framework alongside other methods on GBM datasets, including well-known supervised segmentation backbones for medical images (U-Net~\cite{ronneberger2015u}, DeepLabV3~\cite{chen2017rethinking}, and Swin UNETR~\cite{hatamizadeh2022swin}), the few-shot learning backbone UniverSeg~\cite{valvano2023universeg}, and a zero-shot backbone for GBM segmentation GBMSeg~\cite{yang2024gbmseg}. All methods were implemented using the official settings and hyperparameters to ensure fair comparisons. An ablation study was conducted to assess the impact of different numbers of point prompts for SAM, as well as the effect of varying training epochs on segmentation performance.

\begin{table}[htbp]
  \centering
  \caption{Performance comparison of different segmentation models}
  \label{tab:segmentation_performance}
  \begin{tabular}{lcccc}
    \hline
    Method       & Dice (\%)          & IoU (\%)           & ACC (\%)           & Time (FPS)          \\
    \hline
    U-Net~\cite{ronneberger2015u}       & $17.32 \pm 24.650$  & $12.19 \pm 17.760$  & $89.11 \pm 00.966$ & $14.81 \pm 00.340$  \\
    Swin UNETR~\cite{hatamizadeh2022swin}    & $46.73 \pm 02.658$  & $32.47 \pm 02.796$  & $88.17 \pm 01.052$  & $09.15 \pm 00.194$  \\
    DeepLabV3~\cite{chen2017rethinking}   & $56.30 \pm 12.220$  & $42.54 \pm 10.810$  & $90.20 \pm 04.023$  & $\textbf{15.20} \pm 00.108$   \\
    UniverSeg~\cite{valvano2023universeg}   & $21.46 \pm 07.820$  & $12.80 \pm 05.159$  & $84.77 \pm 02.544$  & $04.87 \pm 00.039$ \\
    GBMSeg~\cite{yang2024gbmseg}      & $68.70 \pm 09.331$  & $56.09 \pm 09.295$  & $93.31 \pm 01.569$  & $00.18 \pm 00.008$\\
    \textbf{CoFi (Ours)}   & $\textbf{74.54} \pm 01.062$  & $\textbf{61.87} \pm 01.217$  & $\textbf{94.26} \pm 00.195$ & $01.90 \pm 00.079$   \\
    \hline
  \end{tabular}
\end{table}

\subsection{GBM Segmentation Performance}
\begin{figure}[htbp]
  \centering
  \includegraphics[width=1.033\textwidth]{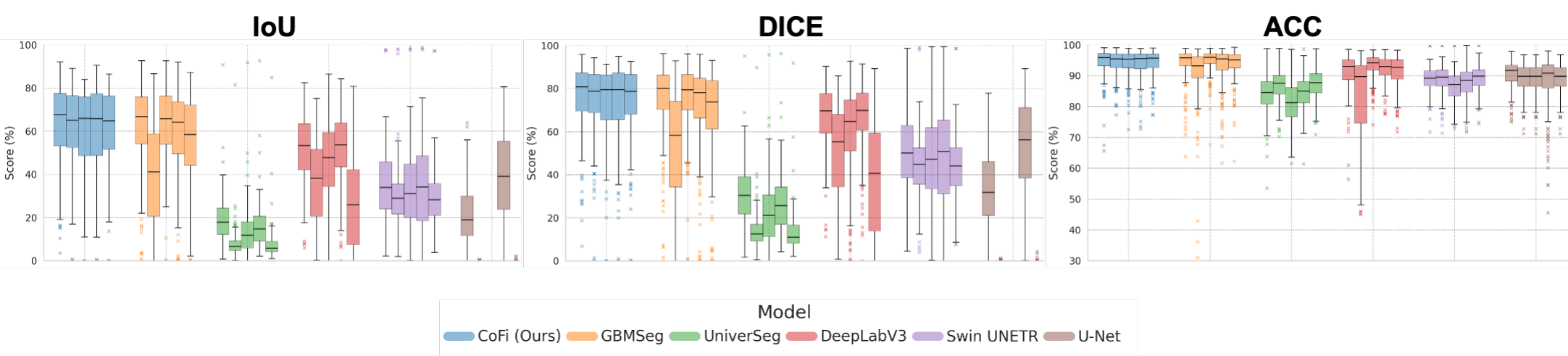}
  \caption{Boxplots of Dice, IoU, and ACC (\%) for six segmentation models (CoFi (Ours), GBMSeg, U-Net, UniverSeg, Swin UNETR, DeepLabV3) across five runs. Boxes span the interquartile range, whiskers extend to 1.5 $\times$ IQR, the line marks the median, and “$\times$” denotes outliers.}
  \label{fig:boxplot}
\end{figure}

Figure~\ref{fig:qualitative_comp} visually illustrates the segmentation results of different models, with CoFi achieving the highest accuracy. Figure~\ref{fig:boxplot} demonstrates that CoFi delivers the most consistent and robust performance across five randomly sampled support sets, particularly in the second set, where all other models exhibit a notable performance drop. Table~\ref{tab:segmentation_performance} presents the numerical segmentation results for all models. Notably, our proposed CoFi framework achieved a mean DSC of 74.54\%, an IoU of 61.87\%, and an ACC of 94.47\% across five independent runs, underscoring its ability to deliver highly accurate GBM segmentations with minimal supervision. The hybrid approach, integrating DeepLabV3 for coarse mask prediction with SAM-based refinement, outperformed DeepLabV3, GBMSeg, Swin-UNETR, U-Net (which occasionally predicted only background), and UniverSeg in segmentation accuracy(Dice and IoU). CoFi achieved a Dice score 5.84\% higher than GBMSeg and operated about ten times faster, surpassing the current state-of-the-art training-free few-shot model. This improvement is evident both quantitatively, through standard segmentation metrics, and qualitatively, as illustrated by visual examples in Figure~\ref{fig:qualitative_comp}.

\begin{figure}[htbp]
  \centering
  \includegraphics[width=0.4\textwidth]{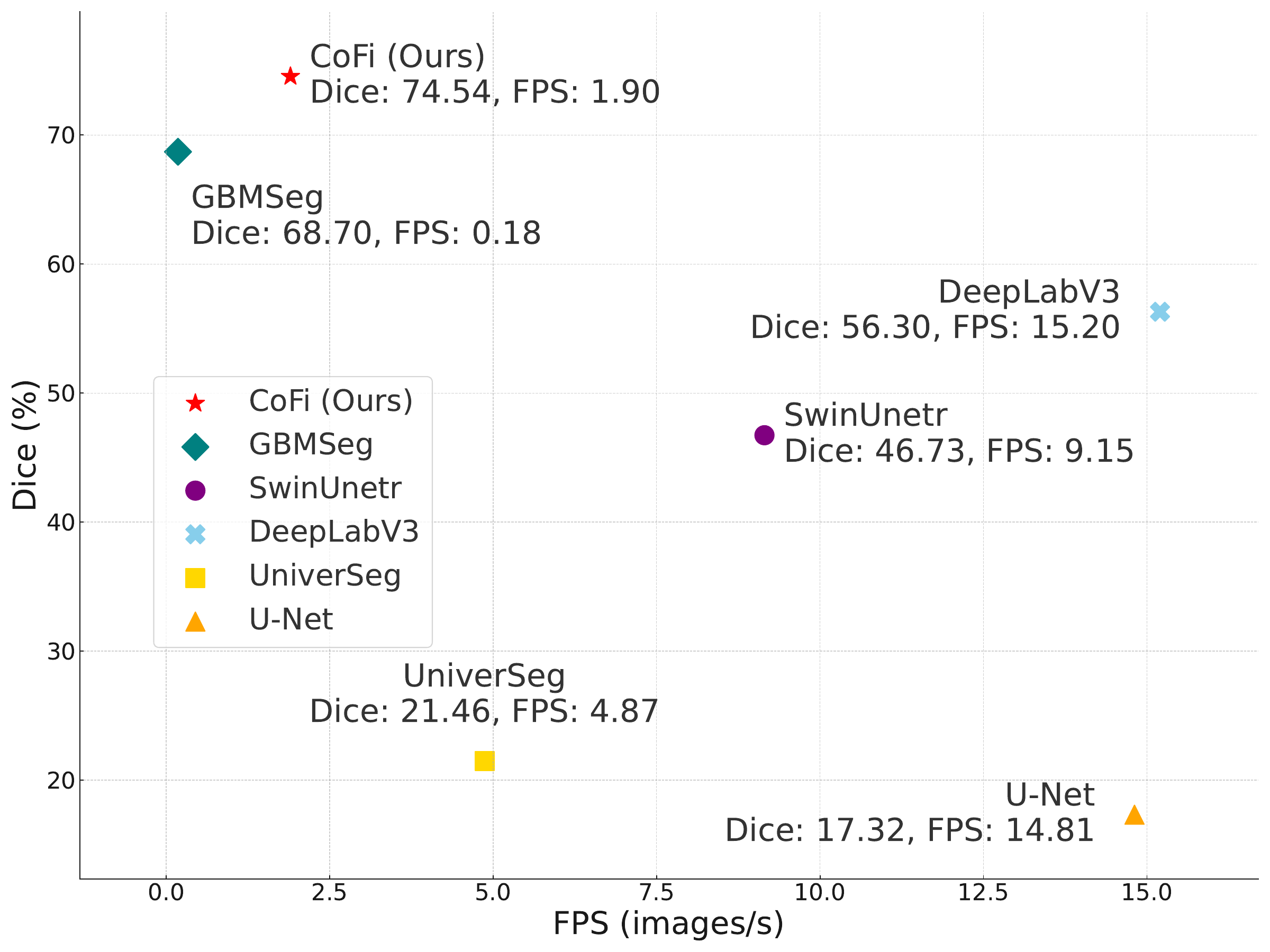}
  \caption{Dice accuracy versus inference speed comparison across models. CoFi achieves the best accuracy while remaining faster than GBMSeg.}
  \label{fig:fps_graph}
\end{figure}

In addition to accuracy, Figure~\ref{fig:fps_graph} and Table~\ref{tab:segmentation_performance} also highlight the computational efficiency of the compared approaches. The results demonstrate that CoFi achieves not only superior segmentation accuracy but also markedly faster inference speeds relative to other few-shot and zero-shot methods. This combination of accuracy and efficiency affirms CoFi’s suitability for practical deployment in both research and time-sensitive clinical environments.

\subsection{Ablation Study}
To further elucidate the factors influencing segmentation performance, we conducted a comprehensive ablation study examining two key aspects: the number of point prompts provided to SAM 2 and the effect of training epochs for the lightweight coarse mask predictor.

\begin{figure}[htbp]
  \centering
  \includegraphics[width=1\textwidth]{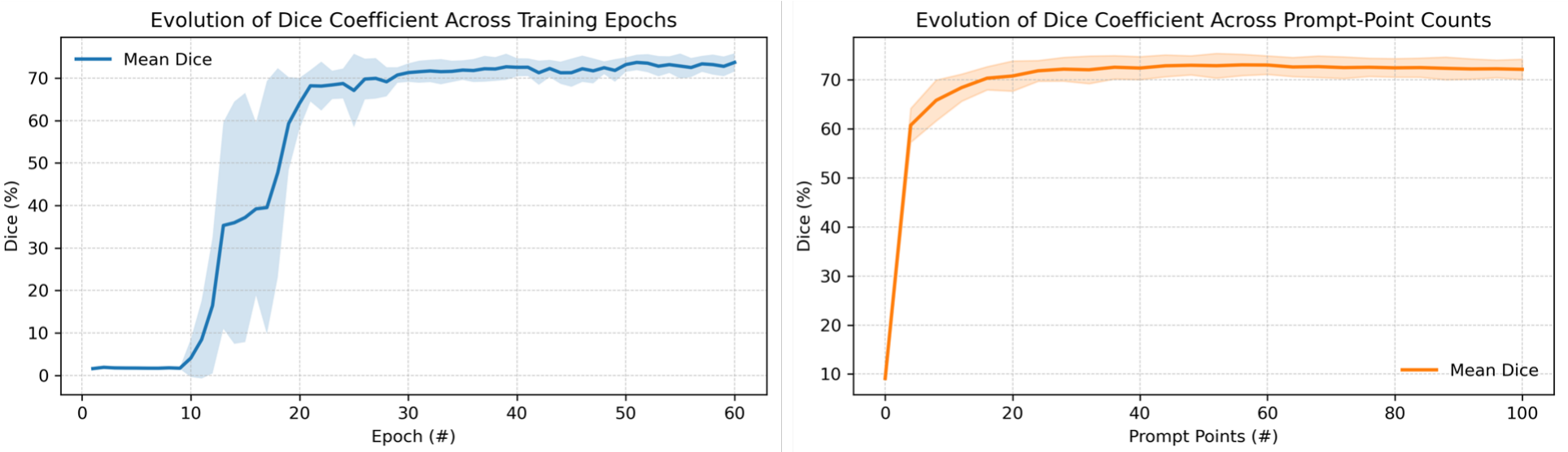}
  \caption{Dice‐score improvement over 5 runs, plotted against training epochs (left) and prompt counts (right); performance stabilizes after approximately 40 epochs and 20 prompts.}
  \label{fig:dice_epoch_point}
\end{figure}

As illustrated in Figure~\ref{fig:dice_epoch_point}, segmentation performance exhibits a marked improvement as the number of point prompts increases, reaching a plateau of consistently high accuracy beyond 40 points (20 positive and 20 negative). This finding demonstrates not only the efficiency of the proposed prompt selection strategy, but also establishes a practical guideline for determining the optimal number of prompts required to maximize segmentation quality without incurring unnecessary computational overhead. Furthermore, Figure~\ref{fig:dice_epoch_point} also highlights the influence of training duration for the lightweight model on the quality of the generated coarse masks. The results reveal that, after 40 training epochs, the model produces stable and reliable point prompts, which in turn enable SAM 2 to generate well-refined segmentation masks.

Collectively, these ablation experiments provide critical insights into optimizing workflow parameters, thereby informing future adjustments to further enhance both the accuracy and efficiency of the CoFi pipeline.

\subsection{Discussion} 
Although CoFi consistently demonstrates strong few-shot segmentation performance, several considerations remain. Segmentation accuracy is influenced by how well the support images represent the structural and textural variability of the dataset; performance remains robust with well-matched support sets but may be modestly affected by pronounced differences in image properties. While the SAM-based refinement stage generally enhances boundary precision, its effectiveness depends on the informativeness of the initial coarse masks. Additionally, the sequential use of DeepLabV3 and SAM 2 may present computational challenges in low-resource environments. Future work will focus on domain adaptation and dynamic support augmentation to further enhance robustness, as well as optimizing computational efficiency to broaden CoFi’s applicability in both clinical and research settings.

\section{new or breakthrough work to be presented}
We present CoFi, a novel pipeline for GBM segmentation in EM that combines few-shot learning with a fast, anatomy-aware prompt generation algorithm. By integrating morphological filtering with efficient point selection, CoFi enables accurate and near real-time segmentation using minimal annotated data, outperforming existing few-shot and zero-shot methods while reducing both annotation and computational demands. This advance offers a practical and scalable solution for robust AI-driven pathology in data-limited clinical settings.

\section{Conclusion} 

In this work, we introduced CoFi, a fast and efficient coarse-to-fine few-shot segmentation framework for accurate delineation of the GBM in grayscale TEM images. By leveraging a lightweight neural network trained on only three annotated images for coarse mask prediction, followed by automated prompt generation and refinement using SAM 2, CoFi demonstrates robust segmentation performance even in challenging low-data scenarios. The pipeline’s unique blend of speed, accuracy, and minimal annotation requirements makes CoFi highly suitable for both research and clinical workflows where rapid and reliable segmentation is essential. Future work will focus on enhancing generalizability across different domains, reducing sensitivity to support set selection, and further optimizing the pipeline for seamless deployment in diverse biomedical imaging settings.

\section{ACKNOWLEDGMENTS} 
This research was supported by the WCM Radiology AIMI Fellowship and WCM CTSC 2026 Pilot Award.

\bibliography{main} 
\bibliographystyle{spiebib} 

\end{document}